\documentclass[letterpaper, 10 pt, conference]{ieeeconf}  

\IEEEoverridecommandlockouts                              

\usepackage{amsmath}
\usepackage{amssymb}
\usepackage[linesnumbered,ruled]{algorithm2e}
\usepackage{mathtools}  
\usepackage{caption}
\usepackage{booktabs}
\usepackage{threeparttable}
\usepackage{kantlipsum}
\usepackage{gensymb}
\usepackage[font=small]{caption}
\usepackage[dvips,frame,rotate,import]{xy} 
\smallbreak
\usepackage[english]{babel}
\usepackage{graphicx}
\graphicspath{}
\usepackage{float}
\usepackage{cite}
\usepackage{multirow}

\title{\Large \bf
T-CBF: Traversability-based Control Barrier Function to Navigate Vertically Challenging Terrain
}

\author{Manas Gupta and Xuesu Xiao
\thanks{All authors are with RobotiXX lab, in the Department of Computer Science, George Mason
University {\tt\scriptsize\{mgupta27, xiao\}}@gmu.edu. This research is supported by National Science Foundation (NSF, 2350352), Army Research Office (ARO, W911NF2320004, W911NF2420027, W911NF2520011), Air Force Research Laboratory (AFRL), US Air Forces Central (AFCENT), Google DeepMind (GDM), Clearpath Robotics, Raytheon Technologies (RTX), Tangenta, Mason Innovation Exchange (MIX), and Walmart.}
}

\begin{document}

\maketitle
\thispagestyle{empty}
\pagestyle{empty}

\begin{abstract}

Safety has been of paramount importance in motion planning and control techniques and is an active area of research in the past few years. Most safety research for mobile robots target at maintaining safety with the notion of collision avoidance. However, safety goes beyond just avoiding collisions, especially when robots have to navigate unstructured, vertically challenging, off-road  terrain, where vehicle rollover and immobilization is as critical as collisions. In this work, we introduce a novel Traversability-based Control Barrier Function (T-CBF), in which we use neural Control Barrier Functions (CBFs) to achieve safety beyond collision avoidance on unstructured vertically challenging terrain by reasoning about new safety aspects in terms of traversability. The neural T-CBF trained on safe and unsafe observations specific to traversability safety is then used to generate safe trajectories. Furthermore, we present experimental results in simulation and on a physical Verti-4 Wheeler (V4W) platform, demonstrating that T-CBF can provide traversability safety while reaching the goal position. T-CBF planner outperforms previously developed planners by 30\% in terms of keeping the robot safe and mobile when navigating on real world vertically challenging terrain.
\\

\end{abstract}

\section{Introduction}
Field robots deployed in real-world applications often need to traverse unstructured, complex, and unpredictable off-road terrain, where safety becomes a top priority. The robotics community has developed numerous techniques to address motion planning safety: Traditional model-based control methods, including Hamilton-Jacobi Reachability Analysis~\cite{bansal2017hamilton}, Control Barrier Functions~\cite{ames2019control}, Control Lyapunov Functions~\cite{freeman1996control}, and Model Predictive Control~\cite{holkar2010overview}, have proven effective but struggles with scalability and generalizability across diverse environments. To overcome these limitations, data-driven approaches~\cite{emam2021data, harms2024neural} are emerging, offering improved adaptability to new environments but at a price of lacking formal safety guarantees.

\begin{figure}
\centering 

   \begin{xy}
    \xyimport(100,100){\includegraphics[width=8.6cm]{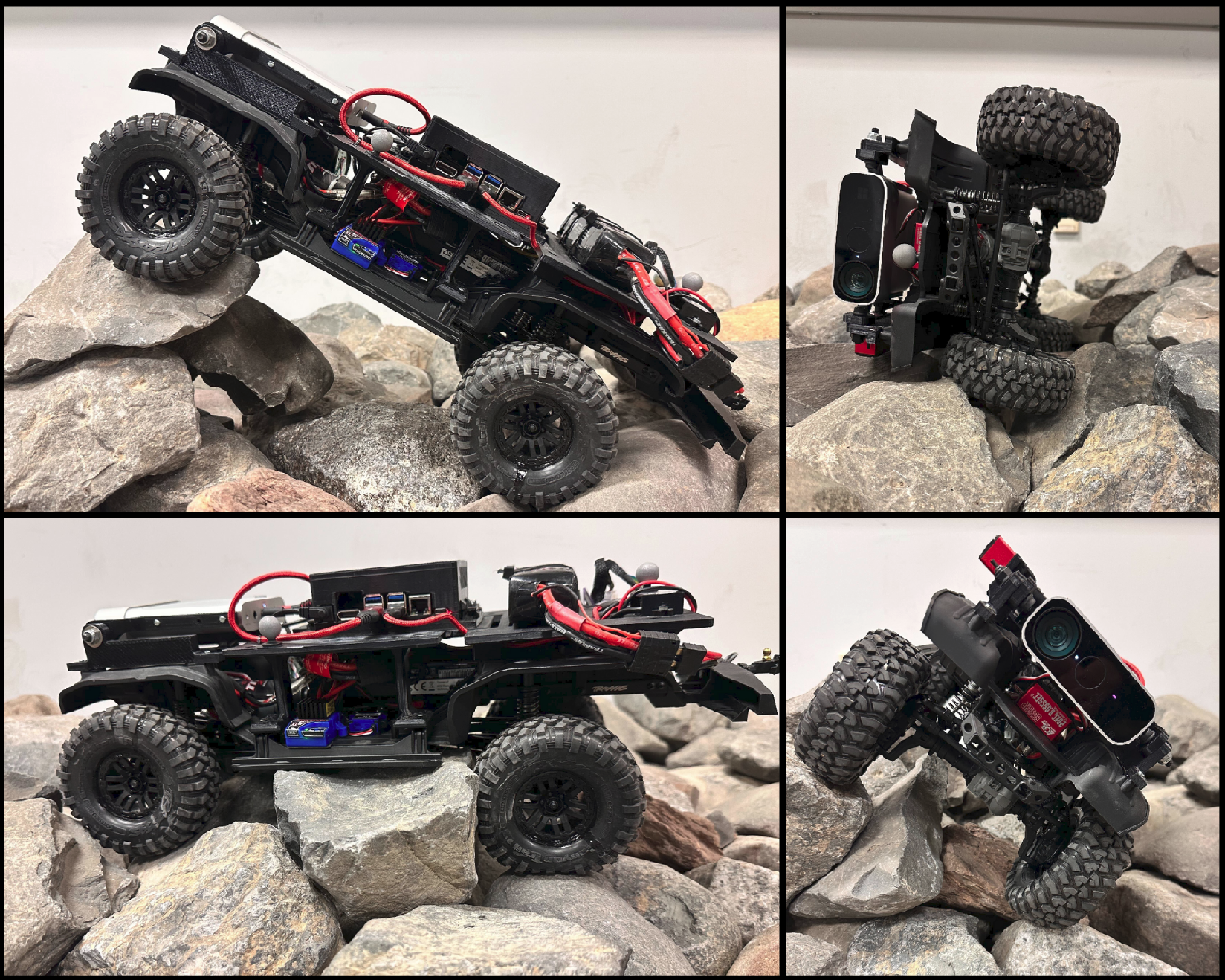}}
   \end{xy} 

        \caption{Unsafe configurations of a wheeled robot navigating vertically challenging terrain. Such risks can be mitigated by a planning algorithm that emphasizes traversability safety beyond traditional collision avoidance.}
  \label{fig:abstract}
      \end{figure}

Despite such a plethora of techniques, most research on mobile robot safety is centered around collision avoidance, defining safety in motion planning as the ability to keep robots within obstacle-free regions. However, in field robotics, safety takes on a more complex meaning, as robots must operate in challenging terrain where the traditional notion of collision avoidance is no longer adequate. Fig.~\ref{fig:abstract} illustrates a wheeled robot navigating unstructured terrain, highlighting scenarios where the robot becomes unsafe when it gets stuck or rolls over in different configurations on vertically challenging terrain~\cite{datar2024toward}. These configurations can be avoided by motion planning algorithms that emphasize on safety beyond collision avoidance, e.g., traversability factors such as vehicle rollover and immobilization. 

Recent advancements in designing scalable and generalizable Control Barrier Functions (CBFs) for mobile robots show promises to facilitate traversability safety beyond collision avoidance. Leveraging machine learning to learn CBFs from data enhances adaptability to diverse environments. To improve generalizability, observation-based approaches have been introduced, offering greater scalability compared to traditional state-based CBFs. These approaches utilize perception inputs, such as LiDAR point clouds or camera images, to extract environmental obstacle information. Existing CBFs in these approaches learn a distance function from the observations to maintain safe distance from obstacles. However, extending the safety notion beyond obstacle avoidance to off-road traversability safety remains a challenge. 

To address this challenge, this work proposes a Traversability-based Control Barrier Function (T-CBF) to enable traversability safety for navigation of wheeled robots on unstructured and vertically challenging terrain using onboard perception inputs. The main contributions are threefolds: (1) The introduction of a novel approach to construct a CBF that extends the concept of safety to traversability by incorporating perception-based information. (2) Demonstrating the generalizability of T-CBF to previously unseen unstructured environments. (3) A statistical evaluation and experimental verification of T-CBF on an autonomous Verti-4 Wheeler (V4W) \cite{datar2024verti} platform in diverse environments. The results demonstrate that T-CBF providesmay prevent the theoretical guarantees outlined in Section V from fully applying in real-world scenarios. To support the practical effectiveness of our learned certificates, the following section presents experimental results demonstrating that the controller consistently operates safely and achieves its objectives across diverse simulated and physical environments. traversability safety beyond collision avoidance while navigating complex, vertically challenging, unstructured environments.

\section{Related Work}\label{sec:relatedWork}
This section discusses the related work on navigation in unstructured environments, the application of CBFs to ensure safety in navigation, and their extension to learning-based approaches. 
\subsection{Navigation in Unstructured Environments}
Navigation for wheeled robots in unstructured off-road environments presents significant challenges. Existing work have investigated various challenges and planning algorithms for navigation in  unstructured off-road environments~\cite{wijayathunga2023challenges, wang2024survey, xiao2022motion}. Due to complexity of modeling system dynamics and terrain interactions, most approaches utilize machine learning to learn kinodynamic model  \cite{xiao2021learning,karnan2022vi,atreya2022high, datar2024terrain, lee2023learning, datar2023learning, sivaprakasam2021improving, cai2025pietra} and terrain representation~\cite{sathyamoorthy2022terrapn, pokhrel2024cahsor, sivaprakasam2021improving, castro2023does} or reduce uncertainty in terrain interactions~\cite{seo2023safe}. However, these methods do not explicitly ensure traversability safety which is crucial for navigation in unstructured terrain. 

\subsection{Control Barrier Functions}
The classical formulation of CBFs provides safety guarantees in various robotics applications. Rabiee and Hoagg~\cite{rabiee2024guaranteed} integrated CBFs into the Model Predictive Path Integral (MPPI) algorithm to ensure safety of sampled trajectories. Similarly, Grandia et al.~\cite{grandia2021multi} applied CBFs within a Model Predictive Control framework to determine safe foot placements for legged robots navigating structured terrain such as stepping stones. For robotic manipulation, Dai et al.~\cite{dai2023safe} employed CBFs for object avoidance using differentiable optimization solvers while maintaining safety guarantees. Furthermore, Patterson et al.~\cite{patterson2024safe} utilized CBFs to regulate self-contact in soft-rigid robots, demonstrating their versatility in maintaining safe interactions across different robotic domains. The classical approaches struggles to generalize to new environments and are often overly conservative in order to provide formal safety guarantees.

\subsection{Learned Control Barrier Functions}
With the growing adoption of data-driven methods, CBFs are becoming increasingly significant in robotics applications. 
In general, synthesizing a CBF requires a mathematical formulation that can either be learned by a neural network or directly incorporated as a constraint in an optimization framework. 
For approaches utilizing perception data, Long et al.~\cite{long2021learning} developed a CBF that is created online using range sensing for obstacle avoidance, and Dawson et al.~\cite{dawson2022learning} designed a hybrid controller incorporating a learned CBF to navigate using LiDAR point cloud data. Similarly, Abidi et al. ~\cite{abdi2023safe} leveraged camera sensor images to construct a vision-based learned CBF for obstacle avoidance while maintaining a safe following distance. Additionally, Harms et al.~\cite{harms2024neural} introduced a neural CBF-based safety filter for quadrotor navigation in unknown environments. These approaches demonstrate the effectiveness of machine learning in designing scalable CBFs that generalize to new environments.
Reinforcement learning has also been used in conjunction with CBFs to develop safe navigation policies for obstacle avoidance~\cite{zhang2021safe, song2022safe, emam2021safe}.

However, all these approaches focus on collision avoidance as a safety concept, which is not sufficient for navigation on unstructured terrain.  
This research extends prior work on CBFs from collision avoidance to traversability safety.

\section{Preliminaries} \label{sec:preliminaries}

Consider a closed-loop non-linear control affine system, 
    \begin{equation}
     \begin{split}
        \dot{\textbf{x}} = f(\textbf{x} ,\textbf{u}), \\
         o  = k(\textbf{x},m) \label{systemeq}
         \end{split}
    \end{equation}
where $ \textbf{x} \in \mathcal{X} \subset \mathbb{R}^n $ and $ \textbf{u} \in \mathcal{U} \subset \mathbb{R}^m $ are the robot state and control input respectively. $o \subset \mathbb{R}^{L*W}$ is the onboard observation,  which is a 2.5D elevation map for vertically challenging terrain with $L$ and $W$ being length and width of the elevation patch. $k(\cdot, \cdot)$ maps the state $\textbf{x}$ and environment variable $m$, which is the terrain underneath the robot, to the 2.5D elevation map observation $o$. $f(x,u)$ is locally Lipschitz continuous function.

\textit{Definition 1:} A set $S \subset \mathbb{R}^n$ is called a forward invariant set for the system \eqref{systemeq}, if for any $\textbf{x}(t_0) \in S$ there exists an input trajectory $\textbf{u}_t \in \mathcal{U}$ such that $\textbf{x}(t) \in S$ $ \forall t \geq t_0$ under system \eqref{systemeq} \cite{blanchini1999set}.

\textit{Definition 2:} (Control Barrier Function) The system \eqref{systemeq} is considered safe if $\textbf{x}(t) \in \mathcal{X}_s \subset \mathcal{X} $, $\textbf{u}_t \in \mathcal{U}_s \subset \mathcal{U}$, $\forall t \geq 0$, where $\mathcal{X}_s$ and $\mathcal{U}_s$ are safe states and safe control inputs respectively.

Let $S \subset \mathcal{X}_s$ be a zero-superlevel set for a smooth continuous differentiable function $h : \mathbb{R}^n \rightarrow \mathbb{R}$ that satisfies the following conditions:
\begin{equation}
    \tag{2a}
     S = \{\textbf{x} \in \mathcal{X} \, | \,  h(\textbf{x}) \geq 0\}
\end{equation}
\begin{equation}
    \tag{2b}
     \partial S = \{\textbf{x} \in \mathcal{X} \, | \,  h(\textbf{x}) = 0\}
\end{equation}
\begin{equation}
    \tag{2c}
     Int(S) = \{\textbf{x} \in \mathcal{X} \, | \,  h(\textbf{x}) > 0\}
\end{equation}
where $\partial S$ is the boundary and $Int(S)$ is the interior of set S respectively. For forward invariant set $S$, the function $h$ with property that $S = \{\textbf{x} \in \mathcal{X} \, | \, h(\textbf{x}) \geq 0\ \}$ and $ \{ \textbf{x} \in \mathcal{X} \, | \, \frac{dh}{d\textbf{x}}(\textbf{x}) = 0 \} \cap \{\textbf{x} \in \mathcal{X} \, | \,  h(\textbf{x}) = 0\} = \emptyset $ is a control barrier function if there exists an extended class $\mathcal{K}_{\infty}$ function $\alpha(\cdot)$ such that for the system \eqref{systemeq} :
\begin{equation}
    \tag{3a}
     h(\textbf{x}) > 0 \: \: \forall \textbf{x} \in S, 
\end{equation}
\begin{equation}
    \tag{3b}
     \sup_{\textbf{u} \in \mathcal{U}}[\mathcal{L}_f h(\textbf{x}) + \mathcal{L}_g h(\textbf{x})\textbf{u}] \geq -\alpha(h(\textbf{x}))
      \label{cbf_eq}
\end{equation}
where  $[\mathcal{L}_f h(\textbf{x}) + \mathcal{L}_g h(\textbf{x})\textbf{u}] = \dot{h}(x,u)$ and $\mathcal{L}_f$ and $\mathcal{L}_g$ are Lie derivatives. Hence, satisfying \eqref{cbf_eq} makes safe set $S$ forward invariant and $\textbf{u}$ a safe control.

\section{Methodology}\label{sec:method}
Based on the preliminaries, we present T-CBF for safe navigation in vertically challenging terrain. 
\subsection{Traversability-based Control Barrier Function}
\begin{figure}
\includegraphics[width=8.7cm, height = 6.0cm]
    {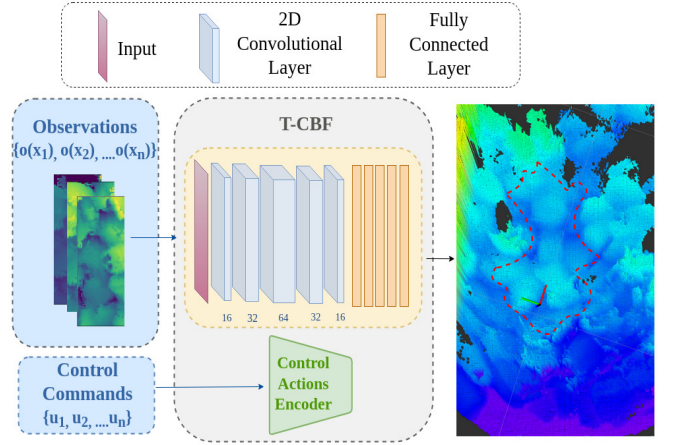}
    \caption{T-CBF Neural Network Architecture: The observations (elevation maps) are processed through a neural network as CBF and control actions are processed through an encoder. The loss function helps learn a CBF enforcing satisfaction of Eqns.~\eqref{hsafe}, \eqref{h_unsafe}, and \eqref{hbf}. The output, safe region (dotted red boundary), is obtained by sampling observations on elevation map from a fixed position with different control commands.}
  \label{fig:T-CBF}
      \end{figure}

Since obstacle avoidance alone is insufficient for safely traversing complex unstructured terrain, we introduce T-CBF, a neural CBF, that is trained on real world observation data and robot dynamic model. The neural network architecture of T-CBF is illustrated in Fig~\ref{fig:T-CBF}. The network takes a 2.5D elevation map patch of size 100 by 40 pixels as input, capturing the elevation information of the terrain. The elevation patch is passed through a series of 2D convolution layers followed by a series of feed-forward layers. The control actions are encoded using an encoder, which is then used in the loss function (described in \ref{sec:4-b}), to train the model. Fig~\ref{fig:T-CBF} illustrates the output of the T-CBF, which is the safe region (dotted red boundary) based on traversability safety.

Formulating safety constraints for high dimensional systems on unstructured terrain is complicated due factors such as suspension and tire deformation, varying tire-terrain friction, vehicle weight distribution and momentum, etc. To address this, we utilize manually driven runs to collect data and identify unsafe states (Fig.~\ref{fig:abstract}) of the robot based on the following criteria: 
Denote the robot state as $\textbf{x}_t = (x_t, y_t, z_t, r_t, p_t, \phi_t)$, where the first three are translational $(x, y, z)$ and last three are rotational (roll, pitch, yaw) components respectively along x, y, and z axis, and control input as $ \textbf{u}_t = (v_t, \omega_t)$, where $v$ is translational velocity and $\omega$ is angular velocity which also correlates to steering curvature.  
\begin{enumerate}
    \item $p_t \geq p_{thresh}$: Pitch angles greater than a threshold value as large pitch angles lead to robot getting stuck; 
    \item $\phi_t \geq \phi_{thresh} $: Roll angles greater than a threshold as large roll angles increases the risk of vehicle rollover;
    \item $ (\Delta x, \Delta y, \Delta z) < \Delta _{thresh}$ and ($ u > u_{thresh}$): Smaller robot position changes with greater velocity and steering commands, compared to their respective thresholds,  indicates that the robot is immobilized, e.g., due to insufficient contact with the terrain or tire slipping.
\end{enumerate}
The first two inequalities can be added as constraints in an optimization framework to maintain safety as those can be predicted by a learned kinodynamic model. However the third inequality is difficult to enforce as a constraint because of the many different factors, as described above, which are involved in robot-terrain interactions after the control is applied. In T-CBF we learn a control barrier function from the samples of safe and unsafe observations which can be applied as a constraint to a control optimization problem.

\subsection{Learning T-CBF from Observations} \label{sec:4-b}
Unlike traditional state-based formulations of CBFs, we define CBF as function of state $\textbf{x}$ and observations
$\textbf{o} \in \mathcal{O} \subset \mathbb{R}^o$ \cite{harms2024neural} \cite{dawson2022learning}. This enables the CBF to generalize into new environments where new observations can be used to create CBF defining safe regions around the robot. The T-CBF based on observations $h : \mathcal{O} \rightarrow \mathbb{R}$ is defined as:

\begin{equation}
    \tag{4a}
    h(o,x) \geq 0 \:\:\:\:\:  \forall o \in \mathcal{O}, \forall x \in \mathcal{X}_\text{safe},\label{hsafe}
\end{equation}
\begin{equation}
    \tag{4b}
       h(o,x) < 0 \:\:\:\:\:  \forall o \in \mathcal{O}, \forall x \in \mathcal{X}_\text{unsafe},\label{h_unsafe}
\end{equation}
\begin{equation}
    \tag{4c}
    h[o(f(x_t,u))] \geq - \alpha_h(h(o,x_t)), \:\:  \forall u \in \mathcal{U}_\text{safe}, \forall x \in \mathcal{X}_\text{safe} \label{hbf}
\end{equation}
where $0 \leq \alpha_h < 1$ is an extended class $\mathcal{K}_{\infty}$ function. We formulate our CBF similar to Harms et al.~\cite{harms2024neural} and use the switching set approach to avoid the observation dynamics approximation that comes from differentiating $h(o,x)$ and results in $\frac{d}{dt}h(o,x) =  h[o(f(x_t,u))]$ which represents the state dynamics of the system.

To train the network illustrated in Fig.~\ref{fig:T-CBF}, we use the following loss function using the elevation map, $o_t$ and $o_{t+1}$ at time $t$ and $t+1$, as an observation input:
\begin{align}
Loss = \: & \tag{5a} \label{eqsafe} c_1 \sum_{\mathclap{(o_t)^i \in \mathcal{O}_\text{unsafe}}} ReLU(h(o_t)^i\texttt{+}\epsilon_1) \:  \texttt{+}&\\ 
     & \tag{5b}\label{equnsafe} c_2 \sum_{\mathclap{(o_t)^i \in \mathcal{O}_\text{safe}}} ReLU(\epsilon_2 \texttt{-} h(o_t)^i) \: \texttt{+} &\\
     &\tag{5c} \label{eqcbf} c_3 \sum_{\mathclap{(o_t, o_{t+1})^i  \in \mathcal{O}_\text{safe}}} ReLU(\epsilon_3 \texttt{-} (h(o_{t+1})^iu_e \texttt{+} \alpha(h(o_t)^i))).
\end{align}
 In the loss function, the terms \eqref{eqsafe} and \eqref{equnsafe} ensure satisfaction of safe and unsafe observation inequalities in \eqref{hsafe} and \eqref{h_unsafe} respectively, which is a standard practice in training a neural CBF. However, we add \eqref{eqcbf} explicitly to enforce satisfaction of equation \eqref{hbf}. In unstructured environment, a terrain patch can be safe to traverse with a control command $u$, but unlike collision avoidance, the same patch can also become unsafe with another control command $u^{\prime}$, hence we explicitly incorporate encoded control command $u_e$ in the equation \eqref{eqcbf} of the loss function  to formulate traversability safety.
 Thresholds $\epsilon_1, \epsilon_2, \epsilon_3 \geq 0$ are added to enforce strict satisfaction of all the inequalities and weights $c_1, c_2, c_3 \geq 0$ are hyper-parameters. 
 
 The learning process involves training the T-CBF network using a small data set of over 4,000 real-world observation samples, evenly distributed between safe and unsafe sets with 20\% reserved for validation. We train the network for 150 epochs using the Adam Optimizer.

\subsection{Planning Algorithm}
We formalize the navigation planning problem as a control optimization framework:
\begin{equation}
    \tag{6a}
    u =  \text{arg} \min_u  \:  \: \lambda_1 \lVert u \rVert   \: \texttt{+} \lambda_2 \:\mathcal{C}_\text{goal} \texttt{+} \lambda_3 \:\mathcal{C}_\text{stab}, \:\label{objective}
\end{equation}
\begin{equation}
    \tag{6b}
       \text{s.t.}  \: \: \: \: \: u_\text{min} \leq u \leq u_\text{max}, \label{system_cons}
\end{equation}
\begin{equation}
    \tag{6c}
    \: \: \: \: \: h_{t+1} + \alpha(h_t) \geq 0.\label{cbf_cons}
\end{equation}
The resulting trajectory adheres to both the system constraint \eqref{system_cons} and the CBF constraint \eqref{cbf_cons} making it traversability safe. To encourage goal-directed behavior, we incorporate a cost term, $\mathcal{C}_\text{goal}$, which penalizes control actions leading to states that deviate from the goal position as shown in equation \eqref{goal_cost}. $x_g$ and $y_g$ are x and y component of goal position and $w_x$ and $w_y$ are weights for x and y components respectively.
\begin{equation}
    \tag{7}
    \mathcal{C}_\text{goal} = w_x |x_t-x_g| + w_y|y_t -y_g| \label{goal_cost} 
\end{equation}
Additionally, we introduce a stability cost, $\mathcal{C}_\text{stab}$ shown in equation \eqref{stab_cost}, to penalize control actions that result in states with excessive pitch $p_t$ and roll $r_t$ which are weighted by $w_p$ and $w_r$ respectively.

\begin{equation}
    \tag{8}
    \mathcal{C}_\text{stab} = w_r |r_t| + w_p|p_t| \label{stab_cost} 
\end{equation}
Given the challenges of mathematically formulating robot dynamics on unstructured terrain, we employ a learned forward kinodynamic model, TAL~\cite{datar2024terrain}, to predict new states based on control inputs. $\lambda_1, \lambda_2, \lambda_3 $ are weights with $\lambda_1 \gg \lambda_2 \gg \lambda_3$ to give more importance on minimizing the control action. The optimization is run in loop for the planning horizon of 10 steps which provides adequate foresight to ensure safe and stable motion towards the goal. The optimization terminates once the robot reaches within 0.1 m of the goal state.

\section{Experiments} \label{sec:exp}
To validate that T-CBF provides traversiability safety on unstructured vertically challenging terrain and is able to generalize to new environment, we compare the performance with WM-VCT \cite{datar2024verti} and TAL \cite{datar2024terrain}, previously developed planners to solve the same problem. We conduct experiments both in simulation and on a physical robot in real world environment. 

\subsection{Robot and Testbed}
The robot used for the physical experiments is a V4W platform \cite{datar2024toward}, an open-source, 1/10th-scale unmanned ground vehicle. The robot features a low-high gear switch and lockable front and rear differentials, enhancing its mobility on vertically challenging terrain. For perception, we utilize the onboard Microsoft Azure Kinect RGB-D camera to generate elevation map and to run Visual Inertia Odometry\cite{chen2023direct}. Real-time elevation maps are generated using an open-source tool that processes the depth data~\cite{miki} and all onboard computation runs on an NVIDIA Jetson Orin NX.

To support experimentation, we construct a 3.1m × 5m rock testbed with a maximum height of 0.6m. For reference, the V4W is 0.2m high, 0.249m wide, 0.523m long, with a wheelbase of 0.312m. The testbed consists of reconfigurable rocks, enabling flexible data collection and mobility testing across diverse terrain configurations.

We categorize the testbed into three difficulty levels—low, medium, and high—based on elevation changes. In the low-difficulty setting, rocks are evenly arranged to minimize elevation variation, ensuring a continuous surface without gaps. The medium difficulty level introduces increased elevation changes along with small gaps, which may cause the robot to lose traction or become temporarily stuck. At the high-difficulty level, significant elevation variations are introduced, with rock placements designed to challenge the robot’s stability, potentially causing rollovers or immobilization. Additionally, wooden slats and spray foam are incorporated to introduce further variations in terrain complexity.

\begin{figure}
\includegraphics[width=8.6cm]
    {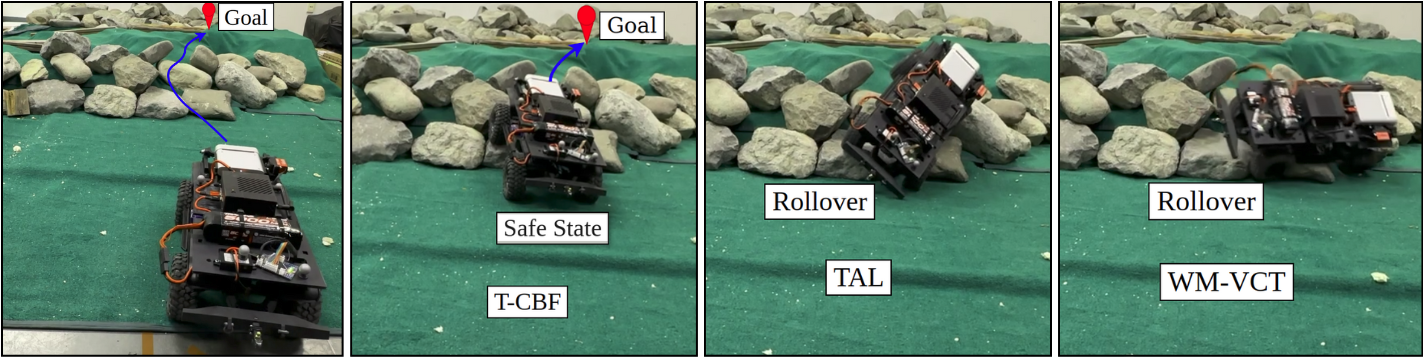}
        \caption{T-CBF validation test (from left to right): V4W receives start and goal (red) states in safe region. We intentionally set a global path that goes through a non-traversable terrain patch (blue). T-CBF navigates the robot towards the goal state and pauses the robot in a safe state. Robot topples and immobilizes with WM-VCT and TAL when attempting to traverse unsafe terrain.   }
  \label{fig:T-CBF_validation}
      \end{figure}

\begin{figure*}
\centering
\includegraphics[width=17.6cm]
    {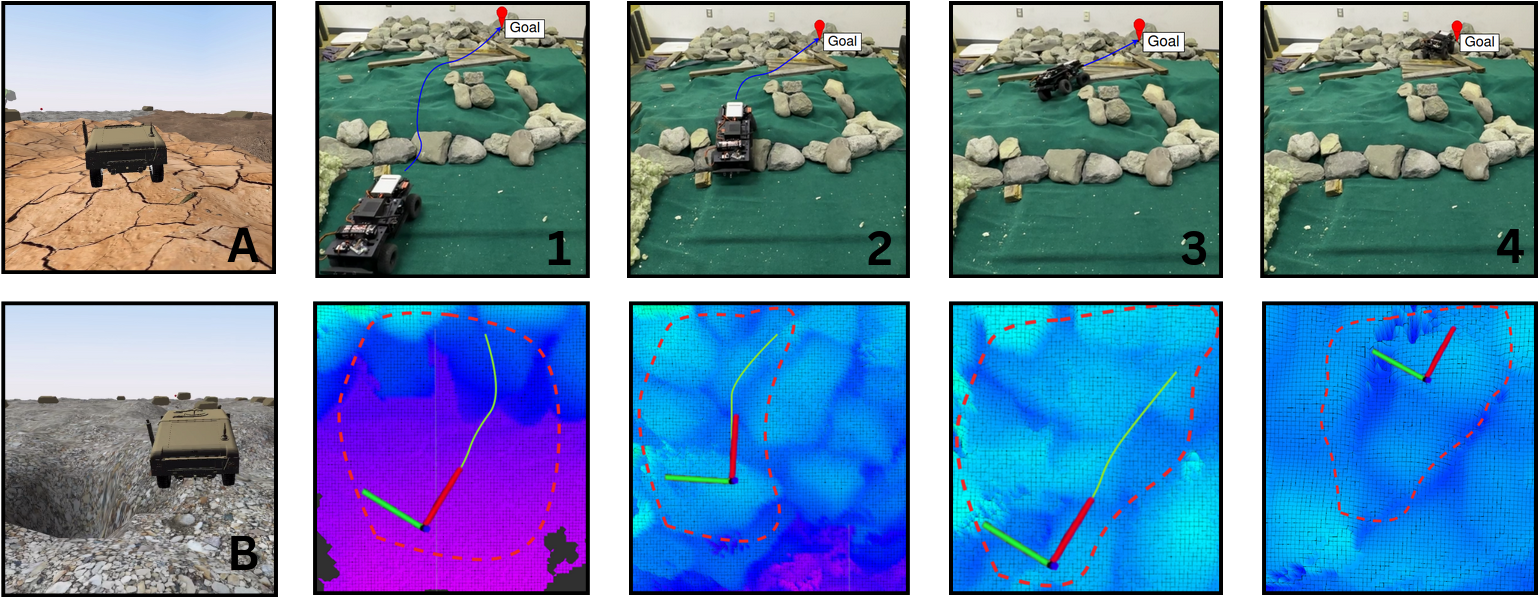}
        \caption{Experimental evaluation of T-CBF. A \& B: Simulation run of robot avoiding unsafe regions while navigating challengin terrain. 1: Robot initialization with start and goal pose in safe region. 2-4: Robot navigating unstructured vertically challenging terrain and avoiding nontraversable unsafe regions to reach the goal position with corresponding bottom images showing traversability-based safe regions (dotted red boundary) on elevation map computed by T-CBF and resulted trajectories in green. }
  \label{fig:final_results}
      \end{figure*}

\vspace{2cm}
\begin{table*}[ht]
    \centering
    \begin{tabular}{llccccc}
        \toprule
        Task & Planner & Success Rate~$\uparrow$  & Traversal Time (s)~$\downarrow$ & Mean Roll (rad)~$\downarrow$ & Mean Pitch (rad)~$\downarrow$ \\
        \midrule
        \multirow{3}{*}{Simulation} 
        & T-CBF  & \textbf{60.7\%} & 26.80 ± 2.30 & \textbf{0.100} ± 0.30 & \textbf{0.073} ± 0.08 \\
        & TAL &  58.2\% & 25.20 ± 2.75 & 0.105 ± 0.35 & 0.076 ± 0.07 \\
        & WM-VCT & 40.12\% & \textbf{24.37} ± 2.52 & 0.154 ± 0.38 & 0.089 ± 0.11 \\
        \midrule

        \multirow{3}{*}{Real World - Low} 
        & T-CBF & \textbf{10/10} & 17.92 ± 2.08 & \textbf{0.131} ± 0.13 & \textbf{0.059} ± 0.08 \\
        & TAL & 10/10 & \textbf{16.63} ± 2.27 & 0.136 ± 0.24 & 0.061 ± 0.07 \\
        & WM-VCT & 9/10 & 17.10 ± 2.10 & 0.154 ± 0.10 & 0.079 ± 0.07 \\
        \midrule

         \multirow{3}{*}{Real World - Medium} 
        & T-CBF & \textbf{10/10} & 18.70 ± 1.10 & \textbf{0.154} ± 0.13 &\textbf{0.087} ± 0.07 \\
        & TAL & 9/10 & \textbf{17.39} ± 2.27 & 0.166 ± 0.24 & 0.091 ± 0.04 \\
        & WM-VCT & 8/10 & 18.16 ± 2.10 & 0.173 ± 0.20 & 0.102 ± 0.06 \\
        \midrule

         \multirow{3}{*}{Real World - High} 
        & T-CBF & \textbf{10/10} & 17.92 ± 1.08 & \textbf{0.176} ± 0.10 & \textbf{0.115} ± 0.03 \\
        & TAL & 7/10 & 16.99 ± 2.27 & 0.189 ± 0.14 & 0.131 ± 0.02 \\
        & WM-VCT & 6/10 & \textbf{14.16} ± 1.10 & 0.197 ± 0.20 & 0.153 ± 0.08 \\
        \bottomrule
    \end{tabular}
    \caption{Performance metrics for various planners across different tasks (best in bold).}
    \label{tab:performance_metrics}
\end{table*}
\vspace{-6em}
\subsection{T-CBF Validation}
To evaluate the safety performance of T-CBF, we configure the testbed to simulate unsafe traversable scenarios. We conduct experiments using the WM-VCT and TAL algorithms, which lack explicit safety constraints, and compare their results with T-CBF. Throughout our experiments, we assume that the robot's start and goal positions are located in a safe region. Unlike the other planners, T-CBF successfully prevents the robot from entering in unsafe states and getting immobilized.  The robot remains mobile and capable of searching alternate paths towards goal position. In contrast, WM-VCT and TAL become immobilized while attempting to navigate the unsafe terrain. The results are presented in Fig.~\ref{fig:T-CBF_validation}. This experiment validates that T-CBF is able to provide traversability safety when navigating unstructured, vertically challenging terrain. While this validation analysis provides useful evidence, it does not constitutes a formal proof. Practical limitations like perception and localization noise and learned neural network architecture may prevent the theoretical grantees in Section IV from fully applying in real world scenarios. To support the practical effectiveness of our learned certificates, the following section presents experimental results demonstrating that the controller consistently operates safely and achieves its objectives across diverse simulated and physical environments.

\subsection{Performance Comparison}
After validating T-CBF's traversability safety beyond collision avoidance, we compare its performance with WM-VCT and TAL in both simulation and real-world experiments. Differential locks of V4W are disabled for the real-world experiments. 

\subsubsection{Simulation}
For simulation we use Verti-Bench~\cite{xu2025verti}, a comprehensive and scalable off-road mobility benchmark designed for extremely rugged, vertically challenging terrain featuring diverse unstructured off-road elements. Built on the high-fidelity multi-physics dynamics simulator Chrono, Verti-Bench captures variations across four orthogonal dimensions, i.e., geometry, semantics, obstacles, and scalability. It provides metrics to compare the performance of different planners on unstructured vertically challenging terrain with different configurations and difficulty levels. We run WM-VCT, TAL, and T-CBF on 30 Verti-Bench environments with terrain difficulty level categorized as low, medium and high. In Each simulation scenario the robot has a specific start and goal position in safe region, and the robot has to navigate the challenging terrain with various surface geometry and semantics to reach the goal location. The success rate in simulation is the percentage of time the robot successfully reaches the goal position.   Table \ref{tab:performance_metrics} shows the overall success rate, traversal time, mean roll, and mean pitch for low, medium, and high difficulty levels combined. T-CBF outperforms other planners in reaching goal location while remaining safe and maintains mobility. It achieves lower roll and pitch angles with the smallest variance, demonstrating T-CBF's superior safety. It is not as aggressive as TAL and WM-VCT and results in longer traversal time. The performance margin against TAL is small because of excess tire slipping in different terrain surface textures and the distribution shift in the observations of elevation map in simulation as training for T-CBF is done on real world data only.

\subsubsection{Real World}
For real-world evaluation, the planners are tested in the rock testbed designed to create varying terrain complexities. The testbed includes reconfigurable rocks to adjust difficulty levels, spray foam covered with green carpet, and wooden slats to introduce surface variations. Each planner is executed 10 times at each of three difficulty levels, resulting in a total of 90 physical runs. The success criteria in real world testing is defined as the ability to keep robot safe, i.e., the robot can move around by executing control actions and searching for feasible paths towards the goal position. Table \ref{tab:performance_metrics} summarizes the experimental results for the three planners on low, medium and high difficulty levels. In the low difficulty environment all planners work well. However, as the complexity of the environment increases, WM-VCT and TAL are unable to generate trajectories to navigate without getting immobilized by either getting stuck or rolling over in many trials. T-CBF outperforms other planners in keeping the robot safe and mobile in all difficult and complex scenarios. It also achieves the smallest and most stable roll and pitch angles. The traversal time of T-CBF planner is longer compared to others because it has to navigate around the unsafe regions to reach goal location.

\section{Conclusions}\label{sec:conclusions}

This work introduces a novel approach for leveraging learned Control Barrier Functions (CBFs) to enable wheeled robots to safely navigate unstructured, vertically challenging terrain. Unlike traditional safety measures that primarily focuse on collision avoidance, the proposed T-CBF framework learns traversability safety from a limited dataset of manually collected runs. Experimental results demonstrate and validate the effectiveness of T-CBF in maintaining both safety and mobility across diverse, vertically challenging environments, showcasing its adaptability to varying terrain configurations and complexities. Given the significant impact of terrain interactions on navigation in unstructured and vertically challenging terrain, incorporating multiple perception inputs can enhance situational awareness and traversability safety. In future work, we aim to extend this framework by integrating observations from camera sensors to enrich terrain perception and further increasing the robustness of T-CBF for field robotics applications.



\begin{thebibliography}{10}
\providecommand{\url}[1]{#1}
\csname url@samestyle\endcsname
\providecommand{\newblock}{\relax}
\providecommand{\bibinfo}[2]{#2}
\providecommand{\BIBentrySTDinterwordspacing}{\spaceskip=0pt\relax}
\providecommand{\BIBentryALTinterwordstretchfactor}{4}
\providecommand{\BIBentryALTinterwordspacing}{\spaceskip=\fontdimen2\font plus
\BIBentryALTinterwordstretchfactor\fontdimen3\font minus \fontdimen4\font\relax}
\providecommand{\BIBforeignlanguage}[2]{{%
\expandafter\ifx\csname l@#1\endcsname\relax
\typeout{** WARNING: IEEEtran.bst: No hyphenation pattern has been}%
\typeout{** loaded for the language `#1'. Using the pattern for}%
\typeout{** the default language instead.}%
\else
\language=\csname l@#1\endcsname
\fi
#2}}
\providecommand{\BIBdecl}{\relax}
\BIBdecl

\bibitem{bansal2017hamilton}
S.~Bansal, M.~Chen, S.~Herbert, and C.~J. Tomlin, ``Hamilton-jacobi reachability: A brief overview and recent advances,'' in \emph{2017 IEEE 56th Annual Conference on Decision and Control (CDC)}.\hskip 1em plus 0.5em minus 0.4em\relax IEEE, 2017, pp. 2242--2253.

\bibitem{ames2019control}
A.~D. Ames, S.~Coogan, M.~Egerstedt, G.~Notomista, K.~Sreenath, and P.~Tabuada, ``Control barrier functions: Theory and applications,'' in \emph{2019 18th European control conference (ECC)}.\hskip 1em plus 0.5em minus 0.4em\relax Ieee, 2019, pp. 3420--3431.

\bibitem{freeman1996control}
R.~A. Freeman and J.~A. Primbs, ``Control lyapunov functions: New ideas from an old source,'' in \emph{Proceedings of 35th IEEE conference on decision and control}, vol.~4.\hskip 1em plus 0.5em minus 0.4em\relax IEEE, 1996, pp. 3926--3931.

\bibitem{holkar2010overview}
K.~Holkar and L.~M. Waghmare, ``An overview of model predictive control,'' \emph{International Journal of control and automation}, vol.~3, no.~4, pp. 47--63, 2010.

\bibitem{emam2021data}
Y.~Emam, P.~Glotfelter, S.~Wilson, G.~Notomista, and M.~Egerstedt, ``Data-driven robust barrier functions for safe, long-term operation,'' \emph{IEEE transactions on robotics}, vol.~38, no.~3, pp. 1671--1685, 2021.

\bibitem{harms2024neural}
M.~Harms, M.~Kulkarni, N.~Khedekar, M.~Jacquet, and K.~Alexis, ``Neural control barrier functions for safe navigation,'' in \emph{2024 IEEE/RSJ International Conference on Intelligent Robots and Systems (IROS)}.\hskip 1em plus 0.5em minus 0.4em\relax IEEE, 2024, pp. 10\,415--10\,422.

\bibitem{robey2020learning}
A.~Robey, H.~Hu, L.~Lindemann, H.~Zhang, D.~V. Dimarogonas, S.~Tu, and N.~Matni, ``Learning control barrier functions from expert demonstrations,'' in \emph{2020 59th IEEE Conference on Decision and Control (CDC)}.\hskip 1em plus 0.5em minus 0.4em\relax Ieee, 2020, pp. 3717--3724.

\bibitem{marvi2021safe}
Z.~Marvi and B.~Kiumarsi, ``Safe reinforcement learning: A control barrier function optimization approach,'' \emph{International Journal of Robust and Nonlinear Control}, vol.~31, no.~6, pp. 1923--1940, 2021.

\bibitem{emam2022safe}
Y.~Emam, G.~Notomista, P.~Glotfelter, Z.~Kira, and M.~Egerstedt, ``Safe reinforcement learning using robust control barrier functions,'' \emph{IEEE Robotics and Automation Letters}, 2022.

\bibitem{datar2024toward}
A.~Datar, C.~Pan, M.~Nazeri, and X.~Xiao, ``Toward wheeled mobility on vertically challenging terrain: Platforms, datasets, and algorithms,'' in \emph{2024 IEEE International Conference on Robotics and Automation (ICRA)}.\hskip 1em plus 0.5em minus 0.4em\relax IEEE, 2024, pp. 16\,322--16\,329.

\bibitem{datar2024verti}
A.~Datar, C.~Pan, and X.~Xiao, ``Verti-wheelers: Wheeled mobility on vertically challenging terrain,'' in \emph{ICRA 2024 Workshop on Resilient Off-road Autonomy}, 2024.

\bibitem{wijayathunga2023challenges}
L.~Wijayathunga, A.~Rassau, and D.~Chai, ``Challenges and solutions for autonomous ground robot scene understanding and navigation in unstructured outdoor environments: A review,'' \emph{Applied Sciences}, vol.~13, no.~17, p. 9877, 2023.

\bibitem{wang2024survey}
N.~Wang, X.~Li, K.~Zhang, J.~Wang, and D.~Xie, ``A survey on path planning for autonomous ground vehicles in unstructured environments,'' \emph{Machines}, vol.~12, no.~1, p.~31, 2024.

\bibitem{xiao2022motion}
X.~Xiao, B.~Liu, G.~Warnell, and P.~Stone, ``Motion planning and control for mobile robot navigation using machine learning: a survey,'' \emph{Autonomous Robots}, vol.~46, no.~5, pp. 569--597, 2022.

\bibitem{xiao2021learning}
X.~Xiao, J.~Biswas, and P.~Stone, ``Learning inverse kinodynamics for accurate high-speed off-road navigation on unstructured terrain,'' \emph{IEEE Robotics and Automation Letters}, vol.~6, no.~3, pp. 6054--6060, 2021.

\bibitem{karnan2022vi}
H.~Karnan, K.~S. Sikand, P.~Atreya, S.~Rabiee, X.~Xiao, G.~Warnell, P.~Stone, and J.~Biswas, ``Vi-ikd: High-speed accurate off-road navigation using learned visual-inertial inverse kinodynamics,'' in \emph{2022 IEEE/RSJ International Conference on Intelligent Robots and Systems (IROS)}.\hskip 1em plus 0.5em minus 0.4em\relax IEEE, 2022, pp. 3294--3301.

\bibitem{atreya2022high}
P.~Atreya, H.~Karnan, K.~S. Sikand, X.~Xiao, S.~Rabiee, and J.~Biswas, ``High-speed accurate robot control using learned forward kinodynamics and non-linear least squares optimization,'' in \emph{2022 IEEE/RSJ International Conference on Intelligent Robots and Systems (IROS)}.\hskip 1em plus 0.5em minus 0.4em\relax IEEE, 2022, pp. 11\,789--11\,795.

\bibitem{datar2024terrain}
A.~Datar, C.~Pan, M.~Nazeri, A.~Pokhrel, and X.~Xiao, ``Terrain-attentive learning for efficient 6-dof kinodynamic modeling on vertically challenging terrain,'' in \emph{2024 IEEE/RSJ International Conference on Intelligent Robots and Systems (IROS)}.\hskip 1em plus 0.5em minus 0.4em\relax IEEE, 2024.

\bibitem{lee2023learning}
H.~Lee, T.~Kim, J.~Mun, and W.~Lee, ``Learning {{Terrain-Aware Kinodynamic Model}} for {{Autonomous Off-Road Rally Driving With Model Predictive Path Integral Control}},'' \emph{IEEE Robotics and Automation Letters}, vol.~8, no.~11, pp. 7663--7670, Nov. 2023.

\bibitem{datar2023learning}
A.~Datar, C.~Pan, and X.~Xiao, ``Learning to model and plan for wheeled mobility on vertically challenging terrain,'' \emph{IEEE Robotics and Automation Letters}, vol.~10, pp. 1505--1512, 2023.

\bibitem{sivaprakasam2021improving}
M.~Sivaprakasam, S.~Triest, W.~Wang, P.~Yin, and S.~Scherer, ``Improving off-road planning techniques with learned costs from physical interactions,'' in \emph{2021 IEEE International Conference on Robotics and Automation (ICRA)}.\hskip 1em plus 0.5em minus 0.4em\relax IEEE, 2021, pp. 4844--4850.

\bibitem{cai2025pietra}
X.~Cai, J.~Queeney, T.~Xu, A.~Datar, C.~Pan, M.~Miller, A.~Flather, P.~R. Osteen, N.~Roy, X.~Xiao \emph{et~al.}, ``Pietra: Physics-informed evidential learning for traversing out-of-distribution terrain,'' \emph{IEEE Robotics and Automation Letters}, 2025.

\bibitem{sathyamoorthy2022terrapn}
A.~J. Sathyamoorthy, K.~Weerakoon, T.~Guan, J.~Liang, and D.~Manocha, ``Terrapn: Unstructured terrain navigation using online self-supervised learning,'' in \emph{2022 IEEE/RSJ International Conference on Intelligent Robots and Systems (IROS)}.\hskip 1em plus 0.5em minus 0.4em\relax IEEE, 2022, pp. 7197--7204.

\bibitem{pokhrel2024cahsor}
A.~Pokhrel, A.~Datar, M.~Nazeri, and X.~Xiao, ``{CAHSOR}: Competence-aware high-speed off-road ground navigation in {SE} (3),'' \emph{IEEE Robotics and Automation Letters}, 2024.

\bibitem{castro2023does}
M.~G. Castro, S.~Triest, W.~Wang, J.~M. Gregory, F.~Sanchez, J.~G. Rogers, and S.~Scherer, ``How does it feel? self-supervised costmap learning for off-road vehicle traversability,'' in \emph{2023 IEEE International Conference on Robotics and Automation (ICRA)}.\hskip 1em plus 0.5em minus 0.4em\relax IEEE, 2023, pp. 931--938.

\bibitem{seo2023safe}
J.~Seo, J.~Mun, and T.~Kim, ``Safe navigation in unstructured environments by minimizing uncertainty in control and perception,'' \emph{arXiv preprint arXiv:2306.14601}, 2023.

\bibitem{rabiee2024guaranteed}
P.~Rabiee and J.~B. Hoagg, ``Guaranteed-safe mppi through composite control barrier functions for efficient sampling in multi-constrained robotic systems,'' \emph{arXiv preprint arXiv:2410.02154}, 2024.

\bibitem{grandia2021multi}
R.~Grandia, A.~J. Taylor, A.~D. Ames, and M.~Hutter, ``Multi-layered safety for legged robots via control barrier functions and model predictive control,'' in \emph{2021 IEEE International Conference on Robotics and Automation (ICRA)}.\hskip 1em plus 0.5em minus 0.4em\relax IEEE, 2021, pp. 8352--8358.

\bibitem{dai2023safe}
B.~Dai, R.~Khorrambakht, P.~Krishnamurthy, V.~Gon{\c{c}}alves, A.~Tzes, and F.~Khorrami, ``Safe navigation and obstacle avoidance using differentiable optimization based control barrier functions,'' \emph{IEEE Robotics and Automation Letters}, vol.~8, no.~9, pp. 5376--5383, 2023.

\bibitem{patterson2024safe}
Z.~J. Patterson, W.~Xiao, E.~Sologuren, and D.~Rus, ``Safe control for soft-rigid robots with self-contact using control barrier functions,'' in \emph{2024 IEEE 7th International Conference on Soft Robotics (RoboSoft)}.\hskip 1em plus 0.5em minus 0.4em\relax IEEE, 2024, pp. 151--156.

\bibitem{xiao2021barriernet}
W.~Xiao, R.~Hasani, X.~Li, and D.~Rus, ``Barriernet: A safety-guaranteed layer for neural networks,'' \emph{arXiv preprint arXiv:2111.11277}, 2021.

\bibitem{long2021learning}
K.~Long, C.~Qian, J.~Cort{\'e}s, and N.~Atanasov, ``Learning barrier functions with memory for robust safe navigation,'' \emph{IEEE Robotics and Automation Letters}, vol.~6, no.~3, pp. 4931--4938, 2021.

\bibitem{dawson2022learning}
C.~Dawson, B.~Lowenkamp, D.~Goff, and C.~Fan, ``Learning safe, generalizable perception-based hybrid control with certificates,'' \emph{IEEE Robotics and Automation Letters}, vol.~7, no.~2, pp. 1904--1911, 2022.

\bibitem{abdi2023safe}
H.~Abdi, G.~Raja, and R.~Ghabcheloo, ``Safe control using vision-based control barrier function (v-cbf),'' in \emph{2023 IEEE International Conference on Robotics and Automation (ICRA)}.\hskip 1em plus 0.5em minus 0.4em\relax IEEE, 2023, pp. 782--788.

\bibitem{zhang2021safe}
L.~Zhang, R.~Zhang, T.~Wu, R.~Weng, M.~Han, and Y.~Zhao, ``Safe reinforcement learning with stability guarantee for motion planning of autonomous vehicles,'' \emph{IEEE transactions on neural networks and learning systems}, vol.~32, no.~12, pp. 5435--5444, 2021.

\bibitem{song2022safe}
L.~Song, L.~Ferderer, and S.~Wu, ``Safe reinforcement learning for lidar-based navigation via control barrier function,'' in \emph{2022 21st IEEE International Conference on Machine Learning and Applications (ICMLA)}, 2022, pp. 264--269.

\bibitem{emam2021safe}
Y.~Emam, P.~Glotfelter, Z.~Kira, and M.~Egerstedt, ``Safe model-based reinforcement learning using robust control barrier functions,'' \emph{arXiv preprint arXiv:2110.05415}, 2021.

\bibitem{blanchini1999set}
F.~Blanchini, ``Set invariance in control,'' \emph{Automatica}, vol.~35, no.~11, pp. 1747--1767, 1999.

\bibitem{chen2023direct}
K.~Chen, R.~Nemiroff, and B.~T. Lopez, ``Direct lidar-inertial odometry: Lightweight lio with continuous-time motion correction,'' in \emph{2023 IEEE international conference on robotics and automation (ICRA)}.\hskip 1em plus 0.5em minus 0.4em\relax IEEE, 2023, pp. 3983--3989.

\bibitem{miki}
T.~Miki, L.~Wellhausen, R.~Grandia, F.~Jenelten, T.~Homberger, and M.~Hutter, ``Elevation mapping for locomotion and navigation using gpu,'' in \emph{2022 IEEE/RSJ International Conference on Intelligent Robots and Systems (IROS)}.\hskip 1em plus 0.5em minus 0.4em\relax IEEE, 2022, pp. 2273--2280.

\bibitem{xu2025verti}
T.~Xu, C.~Pan, M.~B. Rao, A.~Datar, A.~Pokhrel, Y.~Lu, and X.~Xiao, ``Verti-bench: A general and scalable off-road mobility benchmark for vertically challenging terrain,'' \emph{arXiv preprint arXiv:2502.11426}, 2025.

\end{thebibliography}
\end{document}